\pdfoutput=1
\documentclass[10pt, logo, twocolumn, copyright, nonumbering]{nvidiatechreport}
\usepackage{caption}
\usepackage{mdframed}
\usepackage[utf8]{inputenc} 
\usepackage[T1]{fontenc}    
\usepackage{hyperref}       
\usepackage{url}            

\usepackage{booktabs}       
\usepackage{amsfonts}       
\usepackage{nicefrac}       
\usepackage{microtype}      
\usepackage[dvipsnames]{xcolor}         
\usepackage{multirow}
\usepackage{multicol}
\usepackage{graphicx}
\usepackage{subcaption}
\usepackage[numbers]{natbib}
\usepackage{tabto}
\usepackage{xspace}
\usepackage{amsmath}
\usepackage{adjustbox}
\usepackage{enumitem}
\usepackage{wrapfig}
\usepackage{dblfloatfix}
\usepackage{algorithm}
\usepackage{algpseudocode}
\usepackage{comment}
\usepackage{cuted}
\usepackage[most]{tcolorbox}
\usepackage{xcolor}
\usepackage{graphicx}
\usepackage{hyperref}

\newcolumntype{g}{>{\color{gray}}c} 

\title{Nemotron Elastic: Towards Efficient Many-in-One Reasoning LLMs}
\author{
Ali Taghibakhshi*,
Sharath Turuvekere Sreenivas*,
Saurav Muralidharan*,
Ruisi Cai†,
Marcin Chochowski,
Ameya Sunil Mahabaleshwarkar,
Yoshi Suhara,
Oluwatobi Olabiyi,
Daniel Korzekwa,
Mostofa Patwary,
Mohammad Shoeybi,
Jan Kautz,
Bryan Catanzaro,
Ashwath Aithal,
Nima Tajbakhsh,
Pavlo Molchanov
}

\correspondingauthor{X}

\begin{abstract}
\textbf{Abstract:}
Training a family of large language models targeting multiple scales and deployment objectives is prohibitively expensive, requiring separate training runs for each different size. Recent work on model compression through pruning and knowledge distillation has reduced this cost; however, this process still incurs hundreds of billions of tokens worth of training cost per compressed model. In this paper, we present \textbf{Nemotron Elastic}, a framework for building reasoning-oriented LLMs, including hybrid Mamba-Attention architectures, that embed multiple nested submodels within a single parent model, each optimized for different deployment configurations and budgets. Each of these submodels shares weights with the parent model and can be extracted zero-shot during deployment without additional training or fine-tuning. We enable this functionality through an end-to-end trained router, tightly coupled to a two-stage training curriculum designed specifically for reasoning models.
We additionally introduce group-aware SSM elastification that preserves Mamba's structural constraints, heterogeneous MLP elastification, normalized MSE-based layer importance for improved depth selection, and knowledge distillation enabling simultaneous multi-budget optimization. We apply Nemotron Elastic to the Nemotron Nano V2 12B model, simultaneously producing a 9B and a 6B model using only 110B training tokens; this results in over 360× cost reduction compared to training model families from scratch, and around 7x compared to SoTA compression techniques. Each of the nested models performs on par or better than the SoTA in accuracy.
Moreover, unlike other compression methods, the nested capability of our approach allows having a many-in-one reasoning model that has constant deployment memory against the number of models in the family. 
\end{abstract}

\begin{document}
\maketitle
\begin{strip}
\vspace{-30pt} 
\begin{center}
\begin{tcolorbox}[
  colback=gray!3, colframe=gray!30, arc=2pt,
  boxsep=1pt, left=3pt, right=3pt, top=3pt, bottom=3pt,
  width=0.5\textwidth
]
\centering
\textbf{\footnotesize Models on Hugging Face} \\[2pt]  

\raisebox{-0.25\height}{\includegraphics[height=1.2em]{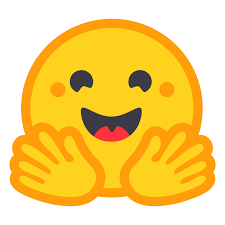}}~%
\href{https://huggingface.co/nvidia/Nemotron-Elastic-12B}{%
  \scalebox{1.2}{\texttt{Nemotron-Elastic}}
}
\end{tcolorbox}
\end{center}
\end{strip}

\section{Introduction}

\begin{figure*}[t]
\centering
\includegraphics[width=0.495\textwidth]{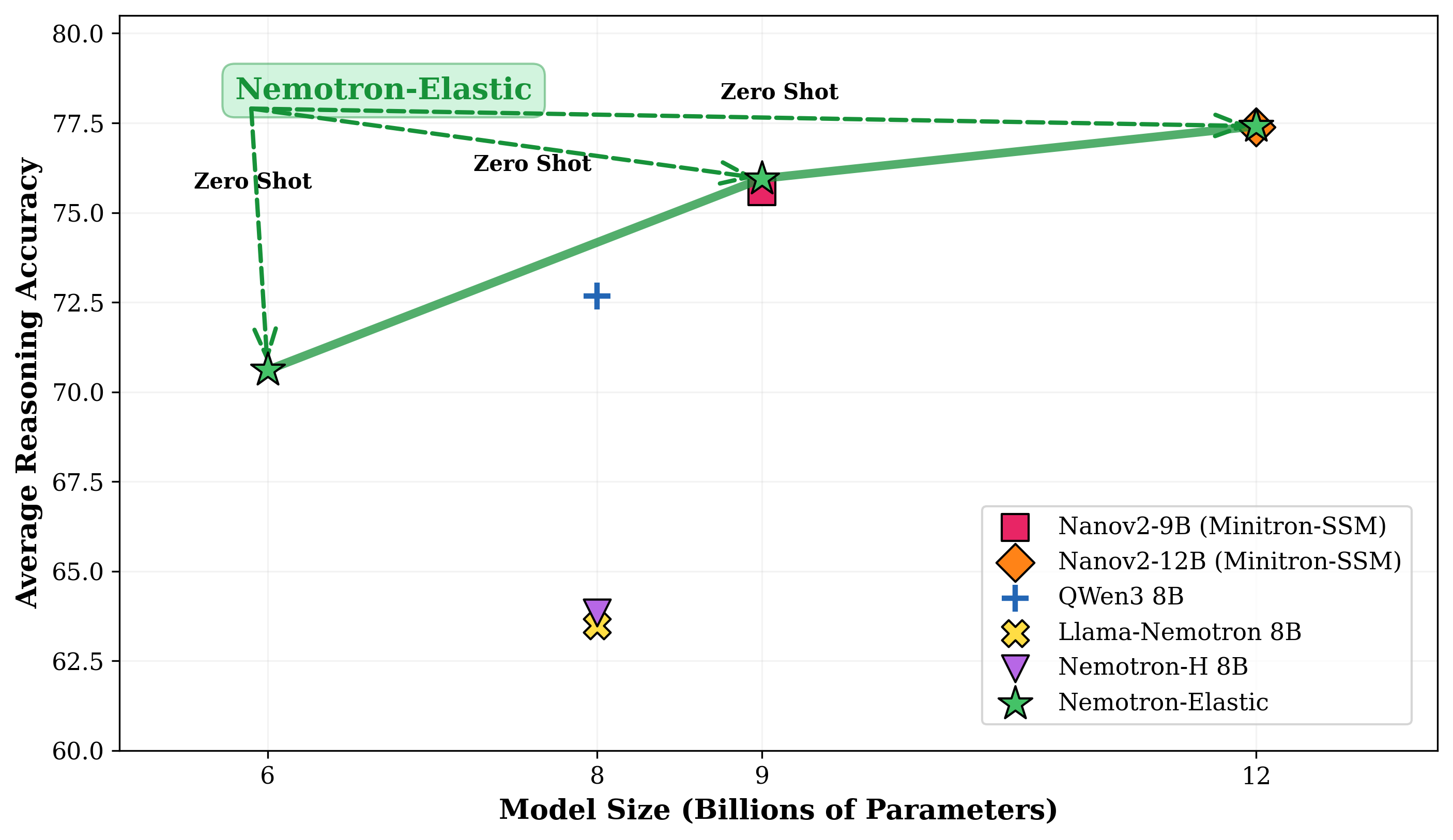}
\hfill
\includegraphics[width=0.495\textwidth]{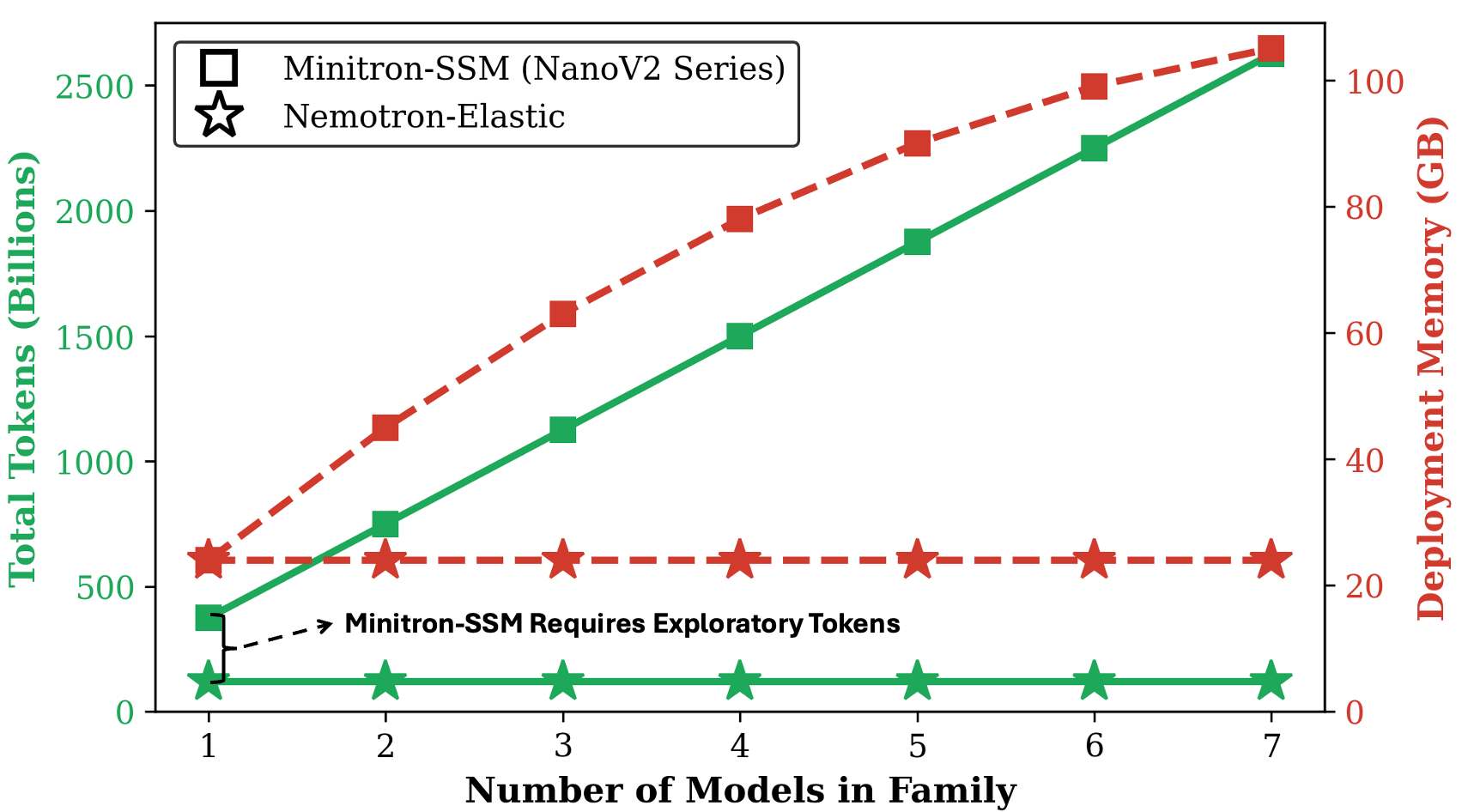}
\caption{
\textbf{Left}: Accuracy across key reasoning and mathematical benchmarks. The accuracy shown is the average across all benchmarks: MATH-500, AIME-2024, AIME-2025, GPQA, LiveCodeBench v5, and MMLU-Pro. 
\textbf{Right}: Scaling analysis comparing Nemotron Elastic and Minitron-SSM as model family size grows. Nemotron Elastic maintains constant cost for tokens and deployment memory, while Minitron-SSM scales linearly.
}
\label{fig:side_by_side_elastic}
\end{figure*}

Large Language Models (LLMs) have demonstrated remarkable capabilities across diverse natural language tasks~\cite{brown2020language, touvron2023llama, dubey2024llama3}, achieving state-of-the-art performance through massive parameter scaling. However, this scaling comes at a significant cost: training LLM families with multiple model sizes—each targeting different deployment scenarios—requires training each variant from scratch, resulting in prohibitively expensive computational budgets. For instance, the Llama-3.1 family~\cite{dubey2024llama3} spans 8B, 70B, and 405B parameters, each trained independently on trillions of tokens. This repeated full-scale training not only multiplies infrastructure costs but also limits practitioners' ability to efficiently deploy models tailored to specific resource constraints.

Recent advances in model compression have sought to address this challenge through structured pruning and knowledge distillation~\cite{muralidharan2024compact, xia2023sheared}. These methods train only the largest model from scratch, then derive smaller variants through pruning and retraining. While effective, they still require hundreds of billions of training tokens per compressed model, keeping overall training costs high. A promising alternative to model compression is elastic or Matryoshka-style nested networks~\cite{cai2024flextron,kudugunta2023matformer}; here, an ``elastic'' or nested model is produced either from scratch or after continued training from an existing model - these elastic models have two special properties: (1) sub-networks meeting specific deployment objectives can be extracted from the parent model ``for free'' (i.e., without any additional training/fine-tuning), and (2) all sub-networks share the same weights with the parent model.

Concurrently, we observe two recent trends that are relevant to the above discussion: the first is the rise of hybrid models that combine attention mechanisms with State Space Models (SSMs) such as Mamba~\cite{gu2023mamba, dao2024transformers}. These hybrid architectures, exemplified by models like Jamba~\cite{lieber2024jamba}, Zamba~\cite{glorioso2024zamba}, and Nemotron-H~\cite{blakeman2025nemotron}, achieve superior efficiency through reduced KV cache requirements and linear-time sequence processing while maintaining competitive accuracy.
Unfortunately, there is very limited work targeting the elastification and compression of hybrid models~\cite{shukla2024matmambamatryoshkastatespace}.
Second is the transition from base and instruct-tuned models to {\em reasoning} models.
Modern reasoning-capable LLMs generate extended chains of thought to solve complex problems, requiring substantial token budgets for intermediate reasoning steps. This creates a fundamental tension: reasoning models demand both architectural flexibility to handle variable computational budgets \emph{and} the capacity to process long-context sequences where multi-step inference unfolds. Existing compression techniques fail to address this dual requirement, as they neither support elastic deployment across diverse constraints nor optimize for the long-context reasoning scenarios critical to these models' performance.

In this work, we present \textbf{Nemotron Elastic}, a framework for training hybrid LLMs that simultaneously support multiple deployment configurations via an end-to-end trained router. Our approach produces multiple nested sub-networks at different parameter budgets from a single elastic training run, each optimized for reasoning through a two-stage curriculum prioritizing long-context capability. We demonstrate that reasoning models require fundamentally different elastic training strategies compared to standard LLMs, with extended-context training (49K tokens) critical for multi-step inference. We achieve up to 40× reduction in training tokens compared to training model families from scratch, while enabling simultaneous training of multiple budgets within the memory footprint of the largest model alone. Our framework achieves this efficiency through: (1) importance-based component ranking establishing architecture priority orderings, (2) frozen teacher knowledge distillation enabling joint sub-network optimization, (3) two-stage curriculum balancing router stabilization with reasoning-specific long-context adaptation, and (4) end-to-end router learning ensuring architecture decisions respond to actual task difficulty rather than post-hoc search heuristics.

We validate our approach by training elastic variants of Nemotron NanoV2 12B reasoning model~\cite{nano2025efficient}, producing both homogeneous and heterogeneous 9B configurations plus a 6B variant, all from a single training run. We notice that the resulting nested models achieve competitive or superior accuracy compared to independently trained baselines while delivering significantly faster inference. This work provides an efficient path toward democratizing access to high-performance reasoning models across diverse deployment scenarios.

This paper makes the following key contributions:

\begin{itemize}
\item \textbf{First elastic reasoning model:} we introduce the first elastic architecture specifically designed for reasoning LLMs, incorporating two-stage training with extended-context optimization (49K tokens) critical for multi-step inference. 

\item \textbf{Depth elastification:} We add depth reduction to elastification via iterative layer removal guided by normalized MSE to the full model’s predictions—resulting in more reliable layer ranking than single-shot or perplexity-based methods.

\item \textbf{Knowledge distillation guided elastification:} during elastic training, we treat the non-elastified model as a fixed teacher, guiding compression using teacher-aligned signals rather than CE loss alone. This results in elastified variants that more closely track the behavior of the original model.

\item \textbf{Significant training cost reduction:} Our approach requires only 110B tokens to derive 6B and 9B variants from a 12B parent—a 7× reduction compared to NanoV2 Compression (Minitron-SSM) and 360× more efficient than NanoV2 Pretraining from scratch.

\item \textbf{Memory-efficient multi-budget training:} Elastic training with nested weight-sharing requires memory overhead of only the largest model plus router parameters ($<$2\% additional memory), enabling simultaneous training and deployment of multiple sizes without incurring a linear increase in memory costs.

\item \textbf{Heterogeneous elastification:} Our router-based search enables layer-wise heterogeneous configurations (e.g., varying FFN dimensions across layers), whereas previous elastic methods support only homogeneous configurations. This allows for more granular and potentially more optimal model candidate exploration.
\end{itemize}

\section{Methodology}

\begin{figure*}[t!]
\centering
\includegraphics[width=1.0\textwidth]{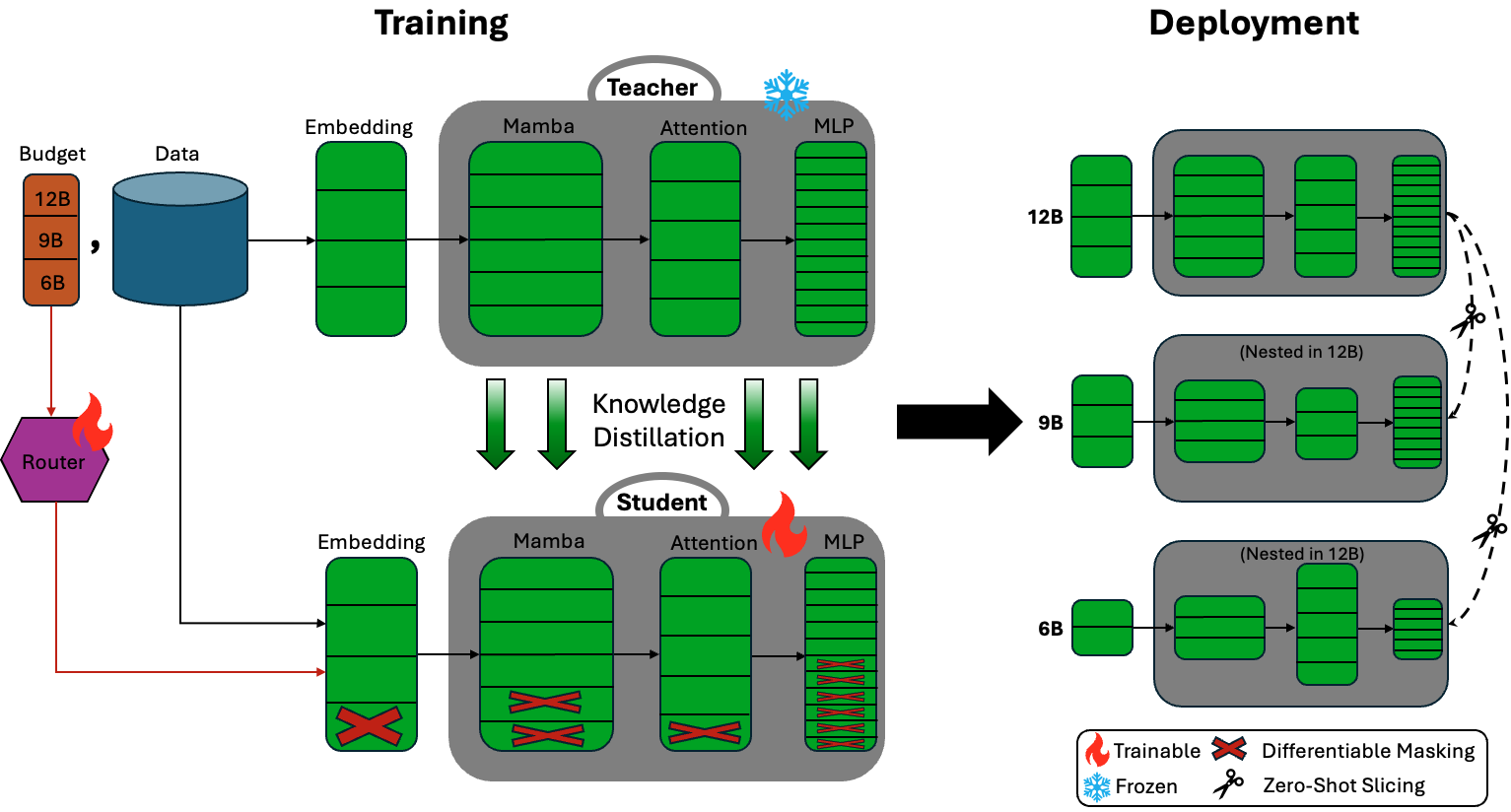}
\caption{
\textbf{Overview of the Nemotron-Elastic training and deployment pipeline.} \textbf{Training:} For each training sample, data flows to both teacher and student models. A budget (parameter size: 6B, 9B, or 12B) is selected and passed to the router, which generates differentiable masks for the student model. Knowledge distillation from the model prior to elastification  enables simultaneous optimization across all budget variants. \textbf{Deployment:} After training, all models are extracted zero-shot from a single elastic checkpoint: the full 12B model and nested sub-networks (9B and 6B) are immediately available without additional fine-tuning or re-training.
}
\label{fig:overview}
\end{figure*}

In this section, we describe the core components of Nemotron Elastic: importance estimation to establish component priority rankings, elastic formulation enabling flexible width and depth selection, two-stage training that couples router learning to task-specific constraints, and the dynamic masking implementation that enables efficient multi-budget training. Figure~\ref{fig:overview} illustrates and overview of the Nemotron Elastic Pipeline.

\subsection{Importance Estimation and Model Preparation}
\label{subsec:importance}
Component importance guides the architectural search by identifying which elements contribute most to model performance. We follow an activation-based approach similar to prior work, establishing a foundation upon which the router makes selection decisions.

\subsubsection{Width}

We employ activation-based importance scoring to rank model components along each width dimension using layer activation magnitudes. For each axis—embedding channels, Mamba heads, Mamba head channels, attention heads, and FFN intermediate neurons—we compute importance scores from forward propagation only, keeping this phase lightweight. 

For embedding channels, we aggregate normalized input activations across the sequence and batch dimensions:
\begin{equation}
\text{Importance}^{(i)}_{\text{emb}} = \sum_{B,L} |\text{LN}(X)|_i
\end{equation}

For FFN neurons, we score based on the output of the first linear layer (the intermediate activations after projection):
\begin{equation}
\text{Importance}^{(i)}_{\text{neuron}} = \sum_{B,L} |X(\mathbf{W}_1^i)^T|
\end{equation}
where \(\mathbf{W}_1^i\) refers to the \(i\)-th row of the first weight matrix in the FFN layer.

For Mamba components, we extract scores from projection matrix outputs (specifically \(\mathbf{W}_x\)) and apply nested procedures that respect group-aware constraints. First, head channels are scored by aggregating across all heads:
\begin{equation}
s_d = \left\|\sum_{B,L} \mathbf{s}_{:,d}\right\|_2
\end{equation}
where \(\mathbf{s} = \text{LN}(X)(\mathbf{W}_x)^T\). Then head-wise scores are computed using top-ranked channels \(\mathcal{D}_{\text{top}}\):
\begin{equation}
f_h = \left\|\mathbf{s}_{h,\mathcal{D}_{\text{top}}}\right\|_2, \quad \forall h \in \{1, \ldots, m_h\}
\end{equation}
Finally, group-constrained ranking is applied to preserve SSM structure, ensuring heads within each Mamba group \(\mathcal{G}_g\) are ranked independently. For attention heads, importance is computed from head-wise activation magnitudes aggregated across query projections. Components are then sorted in decreasing order of importance, establishing a ranking permutation \(\sigma^{(w)}\) that orders components by their contribution to model behavior. This sorted ordering serves as a preference structure guiding the router's selection of which components to retain at different compression budgets.

\subsubsection{Depth}

Layer importance is estimated iteratively using normalized mean squared error (MSE) between the full model's predictions and predictions with specific layers removed. For each layer \(j\), we compute:
\begin{equation}
s_j = \frac{\sum_{B,L} (\mathcal{M}_{\text{full}} - \mathcal{M}_{-j})^2}{\sum_{B,L} \mathcal{M}_{\text{full}}^2}
\end{equation}
where \(\mathcal{M}_{\text{full}}\) represents logits from the full model and \(\mathcal{M}_{-j}\) represents logits with layer \(j\) ablated. The normalization by the full model's energy ensures that importance scores are comparable across different calibration datasets. This yields per-layer importance scores \(\{s_j\}_{j=0}^{N-1}\) that quantify each layer's contribution to model predictions. Layers are sorted in decreasing order of importance, yielding a depth ranking permutation \(\sigma^{(d)}\) that establishes a preference order over the layer stack. This ordering ensures that when the router selects a target depth, the most critical layers—those with highest normalized MSE—are preferentially retained through the binary depth coefficient \(\gamma_j = 1\). This metric-driven approach captures the actual importance structure specific to the model and dataset, enabling principled depth selection during elastic training.

\subsection{Elastic Formulation}

We build upon a nested weight-sharing architecture that enables a single hybrid LLM to dynamically adapt across multiple resource constraints. The model architecture can be resized along both width dimensions (embedding size, attention heads, FFN intermediate dimensions, Mamba heads and head channels) and depth (number of layers), enabling instantaneous generation of sub-networks with different parameter budgets without additional fine-tuning.

\paragraph{Elastic Width.} For width dimensions, we define a set of elastic choices for each component: embedding dimension \(d_e\), FFN intermediate dimension \(d_f\), attention heads \(n_h\), Mamba heads \(m_h\), and Mamba head channels \(m_d\). At training time, sub-networks are constructed by selecting values from these dimension ranges according to a target budget. For a given objective for a sub-network (e.g., latency, memory, model size, etc.), the router selects appropriate dimensions \((d_e^j, d_f^j, n_h^j, m_h^j, m_d^j)\) to satisfy that objective. The nested structure ensures that smaller sub-networks always use a contiguous subset of the neurons, heads, and channels retained by larger variants, achieved through the importance-based ranking established during model preparation. Specifically, embeddings are selected via \(\mathcal{M}_{\text{emb}}\) masking, FFN neurons via \(\mathcal{M}_{\text{ffn}}\), attention heads via \(\mathcal{M}_{\text{attn\_head}}\), and Mamba components via \(\mathcal{M}_{\text{mamba}}\), maintaining consistency with the dynamic masking operators defined in the Implementation section.

\paragraph{Elastic Depth.} Depth elasticity is controlled through a binary selection vector \(\boldsymbol{\gamma}^j = [\gamma^j_0, \gamma^j_1, \ldots, \gamma^j_{N-1}]\) where \(\gamma^j_i \in \{0,1\}\) determines whether layer \(i\) is active in sub-network \(j\). Layers with \(\gamma^j_i = 0\) are bypassed through residual skip connections, maintaining gradient flow while reducing computation. The importance-based layer ranking ensures that critical layers are preferentially retained at lower budgets.

\paragraph{Hybrid Architecture Considerations.} For hybrid models combining Mamba and attention, the elastic formulation must respect the structural constraints of both components. Mamba layers require group-aware pruning and channel consistency to preserve SSM computation and attention layers require head-wise selection. The router jointly optimizes selections across both layer types and all width dimensions to discover architectures that balance the complementary strengths of Mamba's efficient sequence processing and attention's contextual reasoning capabilities.

\subsection{Elastic Training}

\subsubsection{Router Architecture and Design}

For each dynamic dimension \(k \in \{\text{emb}, \text{mamba}, \text{attn\_head}, \text{ffn}, \text{depth}\}\), we introduce a dedicated router network that performs architecture search over the target configuration space. Each router consists of two fully connected layers with leaky ReLU activation applied between them.

\paragraph{Router Input Representation.} The input to router \(k\) is a one-hot encoded vector representing the target compression level:
\begin{equation}
\mathbf{u}^{(k)} = \mathbf{e}_\ell \in \mathbb{R}^{n_{\text{targets}}}
\end{equation}
where \(\mathbf{e}_\ell\) is the \(\ell\)-th standard basis vector and \(n_{\text{targets}}\) is the number of target model configurations.

\paragraph{Router Architecture.} Each router is parameterized as:
\begin{equation}
\mathbf{h}^{(k)} = \text{LeakyReLU}\left(\mathbf{W}^{(k)}_1 \mathbf{u}^{(k)} + \mathbf{b}^{(k)}_1\right)
\end{equation}
where \(\mathbf{W}^{(k)}_1 \in \mathbb{R}^{d_{\text{router}} \times n_{\text{targets}}}\) and \(\mathbf{b}^{(k)}_1 \in \mathbb{R}^{d_{\text{router}}}\) are the first layer weights and bias, and \(d_{\text{router}}\) is the intermediate hidden dimension. The router output is:
\begin{equation}
\mathbf{z}^{(k)} = \mathbf{W}^{(k)}_2 \mathbf{h}^{(k)} + \mathbf{b}^{(k)}_2
\end{equation}
where \(\mathbf{W}^{(k)}_2 \in \mathbb{R}^{n_{\text{out}}^{(k)} \times d_{\text{router}}}\) and \(\mathbf{b}^{(k)}_2 \in \mathbb{R}^{n_{\text{out}}^{(k)}}\). The output dimension \(n_{\text{out}}^{(k)}\) varies by axis and configuration mode.

\paragraph{Router Output Dimensions.} For homogeneous configuration modes where all instances of a component type share the same compression ratio:
\begin{equation}
\begin{split}
n_{\text{out}}^{(\text{emb})} &= |\mathcal{E}| \\
n_{\text{out}}^{(\text{mamba,hom})} &= |\mathcal{M}| \\
n_{\text{out}}^{(\text{attn,hom})} &= |\mathcal{A}| \\
n_{\text{out}}^{(\text{ffn,hom})} &= |\mathcal{F}| \\
n_{\text{out}}^{(\text{depth})} &= N
\end{split}
\end{equation}
where \(|\mathcal{E}|, |\mathcal{M}|, |\mathcal{A}|, |\mathcal{F}|\) denote the cardinality of target configuration sets for each dimension. For heterogeneous configuration modes where each layer can independently select its compression ratio:
\begin{equation}
\begin{split}
n_{\text{out}}^{(\text{mamba,het})} &= |\mathcal{M}| \times N_{\text{M}} \\
n_{\text{out}}^{(\text{attn,het})} &= |\mathcal{A}| \times N_{\text{A}} \\
n_{\text{out}}^{(\text{ffn,het})} &= |\mathcal{F}| \times N_{\text{F}}
\end{split}
\end{equation}
where \(N_{\text{M}}, N_{\text{A}}, N_{\text{F}}\) denote the total counts of Mamba, attention, and FFN layers respectively. Embedding remains homogeneous as its channels are globally indexed.

\subsubsection{Loss Formulation}

The router outputs are passed through Gumbel-Softmax with temperature \(\tau\) to produce soft probability distributions over configuration choices. At each training iteration, we sample from these distributions to obtain relaxed discrete selections that enable gradient flow to the router parameters.

\paragraph{Gumbel-Softmax Relaxation.} Let \(\mathbf{z}^{(k)}\) denote the raw logits output by router \(k\). The Gumbel-Softmax relaxation is:
\begin{equation}
\pi^{(k)}_i = \frac{\exp\left(\frac{\mathbf{z}^{(k)}_i + g_i}{\tau}\right)}{\sum_j \exp\left(\frac{\mathbf{z}^{(k)}_j + g_j}{\tau}\right)}
\end{equation}
where \(g_i \sim \text{Gumbel}(0, 1)\) are i.i.d. Gumbel noise samples and \(\tau > 0\) is a temperature parameter that is annealed from high values (soft exploration) to low values (sharp decisions) during training.

\paragraph{Router Objective Function.} The router is jointly trained to optimize a resource-aware objective that maps selected configurations to hardware and computational constraints. Let \(a_k \in \{1, \ldots, n_{\text{out}}^{(k)}\}\) denote the configuration selected by router \(k\). The resource cost of configuration \(a_k\) is denoted \(\mathcal{C}^{(k)}(a_k)\), where possible cost metrics include parameter count, memory usage (including model parameters, KV cache, Mamba cache, and activations), latency, or throughput. The router loss is:
\begin{equation}
\mathcal{L}_{\text{router}} = \left\| \mathcal{C}^{(k)}(a_k) - \hat{\mathcal{C}}^{(k)} \right\|
\end{equation}
where \(\hat{\mathcal{C}}^{(k)}\) is the target constraint for dimension \(k\). This enables the router to autonomously search through the joint architecture space, balancing multiple objectives and discovering Pareto-optimal configurations. The hybrid approach of combining importance-based sorting with learned router policies is particularly beneficial for hybrid architectures where the interplay between Mamba's linear-time properties and attention's expressiveness creates non-obvious accuracy-efficiency trade-offs.

\subsubsection{Versatile Training Options}

The model and router are jointly optimized during training, enabling the architecture search to directly respond to task-specific learning signals. The model parameters are updated to minimize the primary loss, while the router parameters are updated to discover configurations that satisfy resource constraints while maintaining model accuracy. The training framework supports multiple loss formulations, allowing flexible combinations depending on the training regime and available teacher models.

\paragraph{Cross Entropy Loss} The model can be trained using standard cross-entropy loss over the training corpus without external supervision:
\begin{equation}
\mathcal{L}_{\text{CE}} = -\mathbb{E}_{(x,y) \sim \mathcal{D}}\left[ \log p_\theta(y \mid x) \right]
\end{equation}
where \(\mathcal{D}\) is the training dataset, \(\theta\) represents model parameters, and \(p_\theta(y \mid x)\) is the model's predicted probability distribution. This loss can be used independently or combined with other training objectives.

\paragraph{Knowledge Distillation} Knowledge Distillation (KD) improves model accuracy by transferring knowledge from a teacher model. Let \(p_\theta(x; \tau)\) denote the student model's softmax-normalized logits at temperature \(\tau\), and \(p_\phi(x; \tau)\) denote the teacher's corresponding distribution. The distillation loss using forward KL divergence is:
\begin{equation}
\mathcal{L}_{\text{KD}} = D_{\text{KL}}\left( p_\phi(x; \tau) \| p_\theta(x; \tau) \right)
\end{equation}

\emph{Trainable Teacher}: In this mode, the full-budget model (100\% across all dimensions) simultaneously serves as the teacher and is updated during training. Both student and teacher parameters are optimized jointly:
\begin{equation}
\mathcal{L}_{\text{teacher}} = \mathcal{L}_{\text{KD}}(\theta_{\text{student}}, \theta_{\text{teacher}}) + \alpha \cdot \mathcal{L}_{\text{CE}}(\theta_{\text{teacher}})
\end{equation}
where \(\theta_{\text{teacher}}\) corresponds to model parameters with full budget allocation and \(\alpha > 0\) is a weighting factor. This enables the teacher to adapt to the training distribution while providing moving supervision targets. The Cross-Entropy loss is added in this case so that the model doesn't collapse to itself during self distillation.

\emph{Frozen Teacher}: In this mode, the teacher model parameters are frozen throughout training and do not receive gradient updates. The teacher can either be the original pre-trained full model or an alternative model architecture:
\begin{equation}
\mathcal{L}_{\text{frozen}} = \mathcal{L}_{\text{KD}}(\theta_{\text{student}}, \phi_{\text{fixed}})
\end{equation}
where \(\phi_{\text{fixed}}\) are static teacher parameters. This approach reduces computational overhead and provides stable, consistent supervision throughout training.

\paragraph{Mixed Training Modes} The framework supports flexible combinations of these losses. For example, a training run can employ trainable teachers for initial phases (capturing distribution-specific knowledge) and transition to frozen teachers for final stages (stabilizing convergence). Different sub-models (e.g., elastic variants at different compression levels) can simultaneously use different teacher modes and loss combinations, enabling rich multi-objective training scenarios.

\subsubsection{Final Optimization Target}

The joint optimization of the model and router is achieved through a combined objective:
\begin{equation}
\mathcal{L}_{\text{total}} = \mathcal{L}_{\text{task}}(\theta) + \lambda \cdot \mathcal{L}_{\text{router}}(\boldsymbol{\psi})
\end{equation}
where \(\mathcal{L}_{\text{task}}(\theta)\) is the primary learning objective (either cross-entropy, knowledge distillation, or their combination), \(\boldsymbol{\psi}\) denotes router parameters, and \(\lambda > 0\) is a weighting coefficient that balances task accuracy against resource constraints.

The task loss \(\mathcal{L}_{\text{task}}\) directly incorporates the chosen supervision signal—whether from standard language modeling, knowledge distillation from a teacher, or a hybrid of both. Critically, this end-to-end optimization enables the router to make architecture decisions that are aware of the actual training signal (cross-entropy, distillation loss, or combined), rather than optimizing purely for zero-shot proxy metrics in post-hoc search phases. This tight coupling between NAS and the training objective represents a key distinction from prior methods such as Minitron and Minitron-SSM, which decouple architecture search (performed via importance scoring on frozen checkpoints) from the final training objective. Our approach integrates architecture discovery directly into the learning process, allowing the router to dynamically adapt configurations in response to the loss landscape of the chosen training regime.

\subsubsection{Two-Stage Training with Curriculum-Based Sampling}

Multi-budget elastic training requires carefully orchestrated data allocation across budget targets to prevent training imbalance and maintain performance across all sub-networks. This is particularly critical for reasoning models where task complexity demands sophisticated architectural trade-offs.

\paragraph{Multi-Budget Training Mechanics.}
In the multi-budget setting, each training sample is assigned to one of $n_b$ target budgets, and the corresponding router output determines which subset of parameters participates in the forward pass. This requires careful sampling of data across budgets to ensure balanced learning signals for all model variants. The choice of budget distribution directly influences architecture discovery and performance characteristics of the resulting model family.

\paragraph{The Role of Extended-Context Training for Reasoning.} Standard elastic training approaches optimize for general knowledge recovery and parameter efficiency. However, reasoning tasks impose fundamentally different constraints: complex multi-step inference—from mathematical reasoning to code generation—requires substantial token budget for thinking traces and intermediate steps. Short-context training alone is insufficient for developing genuine reasoning capability; the model must adapt its architecture to support extended sequences where reasoning paths unfold. Extended-context training (with sequence length $L_2$) exposes all elastic variants to problems requiring longer inference chains, forcing the router to discover configurations that maintain coherence and performance across extended contexts. This necessity motivated our two-stage approach: Stage 1 establishes foundational architecture patterns, while Stage 2 enforces reasoning-specific constraints on the final elastic configuration.

\paragraph{Stage 1: Uniform Budget Sampling (Short Context).} During the initial short-context phase (sequence length $L_1$, total tokens $T_1$), we employ uniform budget sampling. For $n_b$ target budgets, each training batch receives equal allocation:
\begin{equation}
p_1(b) = \frac{1}{n_b}, \quad \forall b \in \{1, \ldots, n_b\}
\end{equation}
Uniform sampling ensures all sub-networks receive balanced training signal during router stabilization, allowing architecture discovery without budget-specific bias. This allocation establishes diverse architectural patterns before reasoning becomes the dominant bottleneck.

\paragraph{Stage 2: Curriculum-Based Non-Uniform Sampling (Extended Context).} During extended-context training (sequence length $L_2$, total tokens $T_2$), we transition to non-uniform sampling that prioritizes full-budget models. For $n_b$ target budgets with sampling weights $\{\alpha_1, \alpha_2, \ldots, \alpha_{n_b}\}$ normalized to $\sum_{i=1}^{n_b} \alpha_i = 1$:
\begin{equation}
p_2(b) = \alpha_b, \quad \forall b \in \{1, \ldots, n_b\}
\end{equation}
The curriculum-based distribution addresses training imbalance observed empirically: uniform sampling in extended-context causes performance degradation in the full model while smaller budgets improve, indicating gradient competition. Non-uniform weighting biases updates toward full-model performance—critical when the full model serves as teacher in frozen distillation—while still training smaller variants. This approach prioritizes long-context reasoning capability across all sub-networks, with weights typically skewed toward larger budgets to prevent collapse of the largest model.

\paragraph{Training Signal Coupling to Architecture Search.} The two-stage sampling strategy directly couples multi-budget training to the router's architecture discovery process. During Stage 1, uniform sampling encourages exploration of diverse configurations across budgets. During Stage 2, non-uniform sampling provides stronger gradients for the full model, guiding the router toward configurations that preserve reasoning capability on extended contexts. This coupling ensures that architecture decisions evolve in response to the actual difficulty of training tasks at each stage, rather than being independently determined by importance scores alone.

\subsection{Implementation}

The elastic architecture is instantiated through structured masking applied to the hybrid Mamba-Attention-MLP model. Rather than modifying network topology or creating distinct sub-networks, we apply dimension-specific binary masks that dynamically select active components. This masking-based approach enables efficient training of multiple budgets simultaneously while maintaining architectural transparency and enabling straightforward deployment of any sub-network without architectural recompilation.

\subsubsection {Dynamic Model Formulation}

We present a flexible architecture framework for Nemotron Elastic that enables dynamic adjustment of model dimensions during training through a structured masking approach. Our method builds upon the hybrid Mamba-Attention-MLP architecture and extends the elastic training paradigm to support comprehensive width and depth flexibility for hybrid architectures.

A dynamic model is obtained by making the stack of layers dynamic, and then making each layer type dynamic across different dimensions. If the original LLM is defined as $y = \mathcal{L}_0^N(x)$ where $\mathcal{L}_0^j(x) = \mathcal{L}_0^{j-1}(x) + \mathcal{L}^j(\mathcal{L}_0^{j-1}(x))$, a dynamic layer stack is noted as $\mathcal{D} \circ \mathcal{L}_0^N$ where the operator $\mathcal{D}$ is applied to each layer and makes it dynamic. For example:
\begin{equation}
\mathcal{D} \circ \mathcal{L}^j = (\mathcal{D} \circ \mathcal{L}_j) \cdot \gamma_j
\end{equation}
where $\gamma_j \in \{0, 1\}$ controls layer retention (depth adaptation) and $\mathcal{D} \circ \mathcal{L}_j$ represents a dynamic Mamba, Attention, or MLP layer.

The dynamic operator $\mathcal{D}$ applies dimension-specific binary masks $\mathbf{m}$ to the output activations of each layer component, enabling selective feature retention (width adaptation):
\begin{equation}
\mathcal{D}(\mathcal{L}(x)) = \mathcal{L}(x) \odot \mathbf{m}
\end{equation}
where $\odot$ denotes element-wise multiplication and $\mathbf{m} \in \{0,1\}^d$ is a binary mask vector that determines which dimensions remain active. Depth adaptation is controlled through the binary coefficient vector $\boldsymbol{\gamma} = [\gamma_0, \gamma_1, \ldots, \gamma_{N-1}]$, while width adaptation is managed through dimension-specific masks applied within each layer type.

\subsubsection{Dynamic Mamba}

For Mamba-2 components in the hybrid architecture, we apply group-aware masking following permutation-preserving constraints to maintain structural integrity of state-space computations. The elastic Mamba layer applies the dynamic operator to its output:
\begin{equation}
\mathcal{D}(\text{Mamba}_\ell(y)) = \text{Mamba}_\ell(y) \odot \mathbf{m}_{\text{mamba}}
\end{equation}
where $\mathbf{m}_{\text{mamba}} \in \{0,1\}^{d_e}$ is the output mask constructed from dynamic embedding and Mamba-specific constraints.

\paragraph{Dynamic Embedding Mask Operator.} The operator $\mathcal{M}_{\text{emb}}$ applies to any activation or weight matrix with the hidden size $d_e$ as one dimension. For a matrix $\mathbf{W} \in \mathbb{R}^{d_e \times k}$, the masked operation is:
\begin{equation}
\mathcal{M}_{\text{emb}}(\mathbf{W}) = \mathbf{W} \odot (\mathbf{I}_e \otimes \mathbf{1}_k)
\end{equation}
where $\mathbf{I}_e \in \{0,1\}^{d_e}$ with $\mathbf{I}_e[0:i] = 1$ and $\mathbf{I}_e[i+1:d_e] = 0$ for some $i \in [0, d_e]$, and $\otimes$ denotes outer product broadcasting across dimension $k$. For matrices $\mathbf{W} \in \mathbb{R}^{k \times d_e}$, the mask broadcasts similarly: $\mathcal{M}_{\text{emb}}(\mathbf{W}) = \mathbf{W} \odot (\mathbf{1}_k \otimes \mathbf{I}_e)$. This operator is applied to layer normalization outputs and all weight matrices interfacing with the embedding dimension.

\paragraph{Dynamic Mamba Mask Operator.} The operator $\mathcal{M}_{\text{mamba}}$ applies to matrices where dimensions derive from Mamba heads $m_h$ or head channels $m_d$. For a matrix $\mathbf{W} \in \mathbb{R}^{f(m_h, m_d) \times k}$ where $f$ represents a dimension function (typically $f(m_h, m_d) = m_h \cdot m_d$), the masked operation is:
\begin{equation}
\mathcal{M}_{\text{mamba}}(\mathbf{W}) = \mathbf{W} \odot (\mathbf{I}_m \otimes \mathbf{1}_k)
\end{equation}
where $\mathbf{I}_m \in \{0,1\}^{f(m_h, m_d)}$ is constructed to satisfy:
\begin{equation}
\mathbf{I}_m[\phi(h,c)] = \begin{cases}
1 & \text{if } h \leq h^* \text{ and } c \leq c^* \\
0 & \text{otherwise}
\end{cases}
\end{equation}
with $\phi(h,c)$ mapping head $h$ and channel $c$ to flat index, $h^* \in [0, m_h]$ and $c^* \in [0, m_d]$ defining active dimensions. This construction preserves group-aware permutation structure: for heads $h, h' \in \mathcal{G}_g$ belonging to group $g$, $\mathbf{I}_m[\phi(h,\cdot)] = \mathbf{I}_m[\phi(h',\cdot)]$, and maintains head channel consistency: $\mathbf{I}_m[\phi(\cdot,c)]$ is uniform across all heads for each channel $c$.

\paragraph{Forward Pass.} The dynamic Mamba layer processes input through projection matrices following masked layer normalization. First, we apply the embedding mask to the layer norm output:
\begin{equation}
y_{\text{ln}} = \mathcal{M}_{\text{emb}}(\text{LN}(y))
\end{equation}

Then, projections are computed from the masked normalized input:
\begin{equation}
\begin{split}
z &= \mathbf{W}_z \cdot y_{\text{ln}}, \quad x = \mathbf{W}_x \cdot y_{\text{ln}}, \\
B &= \mathbf{W}_B \cdot y_{\text{ln}}, \quad C = \mathbf{W}_C \cdot y_{\text{ln}}, \quad d_t = \mathbf{W}_{dt} \cdot y_{\text{ln}}
\end{split}
\end{equation}
where $\mathbf{W}_z, \mathbf{W}_x \in \mathbb{R}^{(m_h \cdot m_d) \times d_e}$, $\mathbf{W}_B, \mathbf{W}_C \in \mathbb{R}^{(g \cdot d_s) \times d_e}$, and $\mathbf{W}_{dt} \in \mathbb{R}^{m_h \times d_e}$. Here, $d_e$ is the embedding dimension, $m_h$ denotes Mamba heads, $m_d$ is the head channel dimension, $g$ represents the number of Mamba groups, and $d_s$ is the SSM state dimension.

We apply the Mamba-specific mask to $z$, $x$, and $d_t$:
\begin{equation}
\begin{aligned}
z &\leftarrow \mathcal{M}_{\text{mamba}}(z), \\
x &\leftarrow \mathcal{M}_{\text{mamba}}(x), \\
d_t &\leftarrow \mathcal{M}_{\text{mamba}}(d_t)
\end{aligned}
\end{equation}

The intermediate activations $x$, $B$, and $C$ undergo causal convolution:
\begin{equation}
\hat{x} = \text{conv1d}(x), \quad \hat{B} = \text{conv1d}(B), \quad \hat{C} = \text{conv1d}(C)
\end{equation}
where the conv1d operation on $\hat{x}$ implicitly respects the Mamba mask structure.

The selective state-space model update computes:
\begin{equation}
\tilde{y} = \text{SSM}(\hat{x}, \hat{B}, \hat{C}, \mathbf{A}, \mathbf{D}, d_t)
\end{equation}

Followed by gated RMSNorm and output projection:
\begin{equation}
y_{\text{pre}} = \mathbf{W}_O \cdot \text{RMSNorm}(\tilde{y} \odot \text{silu}(z))
\end{equation}
where $\mathbf{W}_O \in \mathbb{R}^{d_e \times (m_h \cdot m_d)}$.

Finally, both dynamic masks are applied to the layer output:
\begin{equation}
y_{\text{out}} = \mathcal{M}_{\text{emb}}(\mathcal{M}_{\text{mamba}}(y_{\text{pre}}))
\end{equation}

The complete Mamba layer output is thus $\mathcal{D}(\text{Mamba}_\ell(y)) = y_{\text{out}}$.

\subsubsection{Dynamic Attention}

For multi-head attention layers in the hybrid architecture, we apply head-wise and embedding dimension masking to control capacity. The elastic attention layer applies the dynamic operator to its output:
\begin{equation}
\mathcal{D}(\text{Attn}_\ell(y)) = \text{Attn}_\ell(y) \odot \mathbf{m}_{\text{attn}}
\end{equation}
where $\mathbf{m}_{\text{attn}} \in \{0,1\}^{d_e}$ is the output mask constructed from dynamic embedding and attention head constraints.

\paragraph{Dynamic Attention Head Mask Operator.} The operator $\mathcal{M}_{\text{attn\_head}}$ applies to matrices where one dimension derives from attention heads $n_h$ or head dimension $d_h$. For a matrix $\mathbf{W} \in \mathbb{R}^{f(n_h, d_h) \times k}$ where $f(n_h, d_h) = n_h \cdot d_h$, the masked operation is:
\begin{equation}
\mathcal{M}_{\text{attn\_head}}(\mathbf{W}) = \mathbf{W} \odot (\mathbf{I}_a \otimes \mathbf{1}_k)
\end{equation}
where $\mathbf{I}_a \in \{0,1\}^{n_h \cdot d_h}$ satisfies:
\begin{equation}
\mathbf{I}_a[\psi(n,d)] = \begin{cases}
1 & \text{if } n \leq n^* \text{ and } d \leq d^* \\
0 & \text{otherwise}
\end{cases}
\end{equation}
with $\psi(n,d)$ mapping head $n$ and head dimension $d$ to flat index, $n^* \in [0, n_h]$ and $d^* \in [0, d_h]$ defining active dimensions.

\paragraph{Forward Pass.} The dynamic attention layer processes input through masked layer normalization:
\begin{equation}
y_{\text{ln}} = \mathcal{M}_{\text{emb}}(\text{LN}(y))
\end{equation}

Projections for query, key, and value are computed as:
\begin{equation}
\begin{split}
\mathbf{Q} &= \mathcal{M}_{\text{attn\_head}}(\mathbf{W}_Q) \cdot y_{\text{ln}}, \\
\mathbf{K} &= \mathcal{M}_{\text{attn\_head}}(\mathbf{W}_K) \cdot y_{\text{ln}}, \\
\mathbf{V} &= \mathcal{M}_{\text{attn\_head}}(\mathbf{W}_V) \cdot y_{\text{ln}}
\end{split}
\end{equation}
where $\mathbf{W}_Q, \mathbf{W}_K, \mathbf{W}_V \in \mathbb{R}^{(n_h \cdot d_h) \times d_e}$, with $n_h$ denoting attention heads and $d_h$ the head dimension.

The attention computation follows:
\begin{equation}
\text{Attn} = \text{softmax}\left(\frac{\mathbf{Q} \mathbf{K}^T}{\sqrt{d_h}}\right) \mathbf{V}
\end{equation}

Followed by output projection:
\begin{equation}
y_{\text{pre}} = \mathcal{M}_{\text{emb}}(\mathbf{W}_O) \cdot \text{Attn}
\end{equation}
where $\mathbf{W}_O \in \mathbb{R}^{d_e \times (n_h \cdot d_h)}$.

Finally, both dynamic masks are applied to the layer output:
\begin{equation}
y_{\text{out}} = \mathcal{M}_{\text{emb}}(\mathcal{M}_{\text{attn\_head}}(y_{\text{pre}}))
\end{equation}

The complete attention layer output is thus $\mathcal{D}(\text{Attn}_\ell(y)) = y_{\text{out}}$.

\subsubsection{Dynamic FFN}

For feed-forward network layers, we apply masking to both embedding and intermediate dimensions. The elastic FFN layer applies the dynamic operator to its output:
\begin{equation}
\mathcal{D}(\text{FFN}_\ell(y)) = \text{FFN}_\ell(y) \odot \mathbf{m}_{\text{ffn}}
\end{equation}
where $\mathbf{m}_{\text{ffn}} \in \{0,1\}^{d_e}$ is the output mask constructed from dynamic embedding and FFN intermediate dimension constraints.

\paragraph{Dynamic FFN Mask Operator.} The operator $\mathcal{M}_{\text{ffn}}$ applies to matrices where one dimension derives from the FFN intermediate dimension $d_{\text{int}}$. For a matrix $\mathbf{W} \in \mathbb{R}^{d_{\text{int}} \times k}$, the masked operation is:
\begin{equation}
\mathcal{M}_{\text{ffn}}(\mathbf{W}) = \mathbf{W} \odot (\mathbf{I}_f \otimes \mathbf{1}_k)
\end{equation}
where $\mathbf{I}_f \in \{0,1\}^{d_{\text{int}}}$ with $\mathbf{I}_f[0:j] = 1$ and $\mathbf{I}_f[j+1:d_{\text{int}}] = 0$ for some $j \in [0, d_{\text{int}}]$. For matrices $\mathbf{W} \in \mathbb{R}^{k \times d_{\text{int}}}$, the mask broadcasts similarly.

\paragraph{Forward Pass.} The dynamic FFN layer processes input through masked layer normalization:
\begin{equation}
y_{\text{ln}} = \mathcal{M}_{\text{emb}}(\text{LN}(y))
\end{equation}

The first linear transformation with dynamic masking:
\begin{equation}
h = \mathcal{M}_{\text{ffn}}(\mathbf{W}_1) \cdot y_{\text{ln}}
\end{equation}
where $\mathbf{W}_1 \in \mathbb{R}^{d_{\text{int}} \times d_e}$ and $d_{\text{int}}$ is the intermediate dimension.

Followed by activation and second linear transformation:
\begin{equation}
y_{\text{pre}} = \mathcal{M}_{\text{emb}}(\mathbf{W}_2) \cdot \sigma(h)
\end{equation}
where $\mathbf{W}_2 \in \mathbb{R}^{d_e \times d_{\text{int}}}$ and $\sigma(\cdot)$ denotes the activation function.

Finally, both dynamic masks are applied to the layer output:
\begin{equation}
y_{\text{out}} = \mathcal{M}_{\text{emb}}(\mathcal{M}_{\text{ffn}}(y_{\text{pre}}))
\end{equation}

The complete FFN layer output is thus $\mathcal{D}(\text{FFN}_\ell(y)) = y_{\text{out}}$.

\paragraph{Depth Adaptation.} Layer-wise depth adaptation is achieved through selective layer retention controlled by $\boldsymbol{\gamma}$. The set of active layers is:
\begin{equation}
\mathcal{A} = \{j \mid \gamma_j = 1, j \in [0, N-1]\}
\end{equation}
where $|\mathcal{A}| = N_{\text{target}}$ specifies the target model depth. Skipped layers are bypassed via residual connections:
\begin{equation}
y_{j+1} = \begin{cases}
y_j + \mathcal{D} \circ \mathcal{L}_j(y_j) & \text{if } \gamma_j = 1 \\
y_j & \text{if } \gamma_j = 0
\end{cases}
\end{equation}
This maintains signal propagation while reducing computation. For hybrid architectures, selective layer retention enables leveraging the complementary strengths of Mamba and attention components at different model scales.

\subsubsection{Mask Generation}

\paragraph{Mask Generation from Router Output.} The router outputs \(\mathbf{z}^{(k)}\) are processed through Gumbel-Softmax to produce relaxed discrete selections. The selected configuration index is determined by \(\hat{a}_k = \arg\max_i \mathbf{\pi}^{(k)}_i\), where \(\mathbf{\pi}^{(k)}\) is the Gumbel-Softmax probability distribution. In homogeneous mode, if dimension \(k\) selects configuration index \(\hat{a}_k\), the corresponding target count is \(c_{\hat{a}_k}\) (e.g., number of active embedding channels, depth, or head counts per layer). The binary mask is then constructed by selecting the top \(c_{\hat{a}_k}\) components according to the importance-based ranking \(\sigma^{(w)}\) or \(\sigma^{(d)}\):
\begin{equation}
\mathbf{I}^{(k)} = \mathbf{I}[\sigma^{(k)}(j) \leq c_{\hat{a}_k}], \quad j = 1, \ldots, \text{size}^{(k)}
\end{equation}

In heterogeneous mode, the router output is reshaped into per-layer selections: \(\mathbf{z}^{(k)}\) is partitioned into \(N_{\text{X}}\) segments of size \(|\mathcal{X}|\), where each segment determines the configuration for one layer. Per-layer masks are constructed similarly, allowing each layer to have distinct compression ratios. For depth selection, if the router outputs \(L_{\text{target}} \in [1, N]\), the top \(L_{\text{target}}\) layers from the importance ranking \(\sigma^{(d)}\) are activated via \(\gamma_j = 1\) for the selected layers.

The generated masks are then applied to the dynamic model operators \(\mathcal{M}_{\text{emb}}, \mathcal{M}_{\text{mamba}}, \mathcal{M}_{\text{attn\_head}}, \mathcal{M}_{\text{ffn}}\), and depth retention coefficients \(\boldsymbol{\gamma}\) as defined in the Dynamic Model Formulation section, enabling the model to dynamically adjust capacity.

\paragraph{Mask Integration Strategies.} The Gumbel-Softmax probabilities provide differentiable signals for router optimization. We support two mask integration modes:

\emph{Mode 1: Hard Selection via Argmax Logits.} The discrete selection is obtained by \(\hat{i}_k = \arg\max_i \pi^{(k)}_i\), and a hard mask is applied using the corresponding logit:
\begin{equation}
\mathbf{I}^{(k)}_{\text{train}} = \mathbf{z}^{(k)}_{\hat{i}_k} \cdot \mathbf{I}_{\hat{i}_k}
\end{equation}
This directly applies the mask from the selected configuration, scaled by its logit magnitude to provide task-relevant gradient signals.

\emph{Mode 2: Soft Masking via Probabilistic Combination.} Alternatively, masks from all candidate configurations are combined proportionally to their probabilities:
\begin{equation}
\mathbf{I}^{(k)}_{\text{train}} = \sum_i \pi^{(k)}_i \cdot \mathbf{I}_i
\end{equation}
During training, this soft mask is applied in the dynamic operators, allowing gradients to flow through all configuration options. At inference time, the discrete mask corresponding to \(\hat{i}_k\) from the argmax mode is used and the logit, $\mathbf{z}^{(k)}_{\hat{i}_k}$ is set to 1.

\subsection{Elastic Model Deployment}

A key advantage of the elastic architecture is the ability to extract multiple model variants from a single trained checkpoint without requiring separate training or fine-tuning. This is achieved through a learned slicing mechanism that leverages the router module trained during the elastic training phase.

After training converges, the router has learned optimal budget-aware decisions for every layer and component (attention heads, Mamba, FFN, embeddings). At deployment time, to extract a model for any target budget \(B\) that was seen during training, we invoke the router with the budget specification. The router's learned decisions are used to determine which components should be pruned from the full model. These components are then permanently removed (sliced out) from the checkpoint, effectively extracting a nested sub-network that corresponds to the desired parameter count.

Formally, given a trained full model with parameter set \(\Theta_{\max}\) and a target budget \(B \in \mathcal{B}\) (where \(\mathcal{B}\) is the set of budgets used during training), the router \(\mathcal{R}\) produces a pruning specification that identifies the parameters to retain. The sliced model parameters are then:
\[
\Theta_B = \{\theta \in \Theta_{\max} : \theta \text{ is retained for budget } B\}
\]
This zero-shot slicing operation is computationally negligible and produces an inference-ready model immediately, with no retraining, fine-tuning, or additional distillation required. Crucially, any budget \(B \in \mathcal{B}\)—whether the largest, smallest, or any intermediate size explored during training—can be deployed directly from the single full-model checkpoint.

The practical benefit is substantial: practitioners need to deploy and maintain only a single full-size model checkpoint, yet at inference time can select any of the trained budget variants on-the-fly without cost. This enables dynamic model selection based on per-request latency or resource constraints. Furthermore, all extracted variants share the same learned representations and architectural decisions, ensuring consistency across the model family and eliminating the need for separate fine-tuning or calibration for each size.

\section{Experiments and Results}

We evaluate Nemotron Elastic by compressing the NVIDIA Nemotron Nano V2 12B hybrid model~\cite{nano2025efficient} across both base and reasoning variants. We simultaneously target two nested models: a 9B, and a 6B model, representing 25\% and 50\% compression, respectively. This multi-target setting showcases the Nemotron-Elasticibility of our elastic framework to serve multiple deployment scenarios from a single trained model.

\subsection{Experimental Setup}

\paragraph{Training data.} All experiments utilize the same compression data blend that was used to train Nemotron NanoV2 9B (both base and reasoning variants)~\cite{nano2025efficient}. This dataset is employed for both importance estimation of network components, and knowledge distillation-based retraining. Using this standardized data blend ensures fair comparison with the Minitron-SSM baseline~\cite{taghibakhshi2025minitron} and maintains consistency across base and reasoning model variants.

\paragraph{Evaluation tasks.}
We evaluate Nemotron-Elastic across a comprehensive suite of downstream reasoning and knowledge benchmarks. For general knowledge and language understanding, we use MMLU-Pro~\cite{wang2024mmlupro} (college-level multiple-choice reasoning) and GPQA~\cite{rein2023gpqa} (graduate-level science questions). For mathematical and algorithmic reasoning, we employ MATH-500~\cite{hendrycks2021math} (pre-calculus through competition-level mathematics), AIME-2024 and AIME-2025~\cite{aime} (American Invitational Mathematics Examination), and LiveCodeBench v5~\cite{jain2024livecodebench} (code generation and problem-solving). All evaluations use pass@1 metrics with reasoning enabled, averaging results over 4 to 16 shots as appropriate for each benchmark. This diverse evaluation set allows us to assess the quality-efficiency tradeoff of Nemotron-Elastic against baseline compression methods.

\paragraph{Nested compression.}
We simultaneously train three nested models (6B, 9B, and 12B) from a single 12B parent architecture using multi-budget elastic compression. The 12B model serves as the frozen teacher model for knowledge distillation, providing stable supervision signals throughout training. As described in the \textit{Two-Stage Training with Curriculum-Based
Sampling} subsection of the \textit{Methodology} section, training proceeds in two stages: an initial short-context phase (sequence length 8192) followed by an extended-context phase (sequence length 49152).

\paragraph{Hyperparameters and training setup.}
For importance estimation (see the \textit{Importance Estimation and Model Preparation} of the \textit{Methodology} section), we process 1024 calibration samples with a sequence length of 8192. Knowledge distillation training is conducted in two phases:

\emph{Phase 1 (Short Context):} Batch size 1536, sequence length 8192, trained for approximately 65B tokens.

\emph{Phase 2 (Extended Context):} Batch size 512, sequence length 49152, trained for approximately 45B tokens.
 
Model parameters are optimized at a learning rate 9e-5, while router parameters are optimized at 1e-2. A 60-step linear learning rate warmup is applied to both. The Gumbel-Softmax temperature \(\tau\) is initialized at 1.0 and annealed to 0.05. The router weight \(\lambda\) is set to 1.0, and the linear scaling coefficient for router logits is initialized at 1.0 and linearly increased to 10.0. The router intermediate hidden dimension is \(d_{\text{router}} = 256\).

\paragraph{Budget sampling strategy.}
For our Nemotron-Elastic family, we target three nested budgets (6B, 9B, 12B). During the short-context phase, we employ uniform budget sampling:
\begin{equation*}
p(\text{budget}) = \frac{1}{3} \quad \text{for each of } \{6\text{B}, 9\text{B}, 12\text{B}\}
\end{equation*}
This ensures that each budget receives an equal training signal, with approximately one-third of each batch assigned to each model variant.

In the extended-context phase, we transition to weighted non-uniform sampling to prevent accuracy degradation in larger models:
\[
p(\text{12B}) = 0.5,\quad
p(\text{9B}) = 0.3,\quad
p(\text{6B}) = 0.2
\]

The non-uniform distribution biases training toward the full-budget model, addressing an observed empirical phenomenon: under uniform sampling in extended-context training, the 12B model's accuracy substantially degrades while the 6B model improves, indicating a training imbalance. The adjusted weighting recovers this balance, allowing all model variants to maintain strong performance.

\subsection{Results}

\begin{table*}[h]
\centering
\small
\setlength{\tabcolsep}{5pt}
\renewcommand{\arraystretch}{1.0}
\begin{tabular}{l|cccccc|c}
\hline
\textbf{Model} & \textbf{Math-500} & \textbf{AIME-2024} & \textbf{AIME-2025} & \textbf{GPQA} & \textbf{LiveCodeBench} & \textbf{MMLU-Pro} & \textbf{Average} \\
\hline
Nemotron-Elastic-6B & 96.50 & 77.64 & 68.13 & 53.78 & 60.95 & 66.65 & 70.61 \\
Nemotron-Elastic-9B & 97.25 & 80.26 & 75.42 & 62.50 & 66.82 & 73.45 & 75.95 \\
Nemotron-Elastic-12B & 97.70 & 83.44 & 75.83 & 63.25 & 68.01 & 76.20 & 77.41 \\
\hline
NanoV2-9B & 97.30 & 80.89 & 71.43 & 63.01 & 67.30 & 73.61 & 75.99 \\
NanoV2-12B & 97.50 & 82.90 & 72.50 & 65.28 & 67.61 & 78.47 & 77.38 \\
\hline
QWen3-8B & 96.3 & 75.83 & 69.31 & 59.61 & 59.5 & 75.50 & 72.68 
\\
\hline
\end{tabular}

\caption{Multi-budget nested compression results on comprehensive reasoning benchmarks. All three Nemotron-Elastic variants (6B, 9B, 12B) are obtained from a single training run with a frozen 12B teacher. Nemotron-Elastic-12B achieves competitive performance (77.41) compared to NanoV2-12B baseline (77.38), while simultaneously enabling efficient 9B and 6B deployments.}
\label{tab:multibench-results}
\end{table*}

Our multi-budget elastic compression strategy yields three model variants from a single training run, each operating at different parameter budgets while sharing a common foundation. As shown in Table~\ref{tab:multibench-results}, the Nemotron-Elastic-12B model achieves performance comparable to NanoV2-12B on most reasoning benchmarks, achieving an average score of 77.41 compared to 77.38 for NanoV2-12B, despite the complexity of simultaneously optimizing three nested budget targets. Notably, the two-stage training approach with adjusted budget sampling prevents accuracy degradation in larger models that would occur under naive uniform sampling. The extended-context phase training (49k sequence length) demonstrates that the router can adapt architecture decisions to support longer contexts while maintaining multi-budget compatibility. The ability to derive three distinct model deployments from a single training process provides significant practical advantages: a unified model infrastructure can serve heterogeneous hardware constraints and latency requirements through dynamic budget selection without retraining or managing multiple checkpoints.

\paragraph{Cost savings.}
As shown in Tables~\ref{tab:token-comparison} and~\ref{tab:memory-comparison}, Nemotron Elastic achieves substantial reductions in both training token requirements and deployment memory compared to prior compression approaches. These savings become increasingly significant as model family size grows, demonstrating the practical advantages of elastic training over sequential compression methods.

\paragraph{Training token efficiency.}
A key advantage of Nemotron Elastic is the elimination of exploratory knowledge distillation runs required by prior methods such as Minitron~\cite{muralidharan2024compact} and Minitron-SSM~\cite{taghibakhshi2025minitron}. These methods perform architecture search by pruning and distilling candidate configurations to identify optimal architectures for each target size, then perform final knowledge distillation with the selected architecture. This two-phase approach incurs substantial token costs that scale linearly with the number of models in the family: each model size requires both exploratory search runs and final distillation.

In contrast, Nemotron Elastic performs end-to-end router-guided architecture search during a single elastic training run, where all target budgets are optimized simultaneously. The router learns to select optimal configurations for each budget as part of the unified training objective, eliminating the need for separate exploratory runs. Table~\ref{tab:token-comparison} compares training token requirements for deriving 6B and 9B models from a 12B parent.

\begin{table*}[h]
\centering
\small
\setlength{\tabcolsep}{4pt}
\renewcommand{\arraystretch}{1.1}
\begin{tabular}{l|l|cc|c}
\hline
\textbf{Method} & \textbf{Model Sizes} & \textbf{Exploratory} & \textbf{Final} & \textbf{Total} \\
\hline
NanoV2 Pretraining & 6B + 9B & 0 B & 40 T & 40 T \\
NanoV2 Compression (Minitron-SSM) & 6B + 9B & 480 B & 270 B & 750 B \\
Nemotron Elastic & 6B + 9B & 0 B & 110 B & \textbf{110 B} \\
\hline
\end{tabular}
\caption{Token budget comparison for deriving 6B and 9B models. Nemotron Elastic eliminates exploratory runs and requires only a single elastic distillation phase, achieving around 7X token reduction compared to Minitron-SSM (NanoV2 Compression). Note: Token budgets for Minitron-SSM 6B, pretraining 9B, and 6B NanoV2 models are estimated based on the token counts for pretraining and compressing the NanoV2 12B model~\cite{nano2025efficient}.}
\label{tab:token-comparison}
\end{table*}

For prior methods like Minitron-SSM, token cost scales as:
\begin{equation}
\text{Tokens}_{\text{Minitron}}(n) = n \cdot (\text{Tokens}_{\text{explore}} + \text{Tokens}_{\text{KD}})
\end{equation}
where $n$ is the number of target model sizes. In contrast, Nemotron Elastic requires:
\begin{equation}
\text{Tokens}_{\text{Elastic}}(n) = \text{Tokens}_{\text{elastic-KD}} \approx \text{constant}
\end{equation}
This constant-cost property stems from simultaneous multi-budget optimization: all nested sub-networks share gradient information and are trained together, with marginal overhead for additional target budgets.

\paragraph{Deployment memory efficiency.}
Elastic models with nested weight-sharing provide significant memory advantages for deployment scenarios requiring multiple model sizes. Since all sub-networks share the same parameter space with only routing metadata differentiating them, deploying all budget variants requires memory equivalent to the largest model alone. In contrast, traditional compression methods produce separate checkpoints for each model size, requiring cumulative storage.

\begin{table}[h]
\centering
\small
\setlength{\tabcolsep}{5pt}
\renewcommand{\arraystretch}{1.1}
\begin{tabular}{l|cc}
\hline
\textbf{Config} & \textbf{Models} & \textbf{Memory} \\
\hline
Nemotron Elastic & 6B + 9B + 12B & \textbf{24 GB} \\
NanoV2 & 9B + 12B & 42 GB \\
\hline
\end{tabular}
\caption{Deployment memory comparison (BF16 weights). Despite storing \emph{three} models, Nemotron Elastic uses 43\% less memory than NanoV2's \emph{two} models.}
\label{tab:memory-comparison}
\end{table}

The memory advantage scales with family size. For prior approaches, memory requirements scale linearly:
\begin{equation}
\text{Memory}_{\text{Separate}}(n) = \sum_{i=1}^n \text{Size}(\text{Model}_i)
\end{equation}
For elastic models with nested weight-sharing:
\begin{equation}
\text{Memory}_{\text{Nested}}(n) = \text{Size}(\text{Model}_{\max}) + \epsilon_{\text{router}}
\end{equation}
where $\epsilon_{\text{router}} < 0.02 \cdot \text{Size}(\text{Model}_{\max})$ represents router parameter overhead (typically $<$1 GB). The nested architecture is particularly valuable for edge deployment scenarios where multiple model sizes must be available to handle varying workloads or user-selected quality-latency tradeoffs.

\subsection{Effects of Two-Stage Training}

The necessity of two-stage training is demonstrated through comparisons between Stage 1 (short-context) and Stage 2 (extended-context) performance:

\begin{table*}[h]
\centering
\small
\setlength{\tabcolsep}{6pt}
\renewcommand{\arraystretch}{1.1}
\begin{tabular}{l|cc|c|c}
\hline
\textbf{Model} & \multicolumn{2}{c|}{\textbf{Performance}} & \textbf{Absolute} & \textbf{Relative} \\
\textbf{(Benchmark)} & \textbf{Stage 1} & \textbf{Stage 2} & \textbf{Gain} & \textbf{Improvement} \\
\hline
Nemotron-Elastic-6B (Math-500) & 95.15 & 96.50 & +1.35 & +1.4\% \\
Nemotron-Elastic-6B (AIME-2025) & 56.88 & 68.13 & +11.25 & +19.8\% \\
Nemotron-Elastic-6B (GPQA) & 49.12 & 53.78 & +4.66 & +9.5\% \\
\hline
Nemotron-Elastic-9B (Math-500) & 97.13 & 97.25 & +0.12 & +0.1\% \\
Nemotron-Elastic-9B (AIME-2025) & 68.75 & 75.42 & +6.67 & +9.7\% \\
Nemotron-Elastic-9B (GPQA) & 59.43 & 62.50 & +3.07 & +5.2\% \\
\hline
Nemotron-Elastic-12B (Math-500) & 97.27 & 97.70 & +0.43 & +0.4\% \\
Nemotron-Elastic-12B (AIME-2025) & 72.92 & 75.83 & +2.91 & +4.0\% \\
Nemotron-Elastic-12B (GPQA) & 62.50 & 63.25 & +0.75 & +1.2\% \\
\hline
\end{tabular}
\caption{Two-stage training improvements across model sizes and benchmarks. Stage 2 (extended-context) provides substantial gains on reasoning benchmarks, particularly on AIME-2025, where smaller models benefit significantly (6B: +19.8\%, 9B: +9.7\%). These improvements demonstrate that reasoning tasks require extended-context training to achieve competitive performance, validating the two-stage approach.}
\label{tab:two-stage-efficacy}
\end{table*}

The results in Table~\ref{tab:two-stage-efficacy} reveal a clear pattern: Stage 2 extended-context training delivers disproportionate improvements on complex reasoning benchmarks (AIME-2025), especially for smaller models. The 6B model gains 19.8\% on AIME-2025, while the 12B model gains 4.0\%, indicating that smaller models particularly benefit from extended-context adaptation for multi-step reasoning. These gains justify the two-stage curriculum: short-context training stabilizes the router and helps initial recovery of the compressed sub-models, while extended-context improves the long context reasoning capability of the model, necessary for achieving competitive results on reasoning benchmarks.

We evaluate the impact of sampling strategy on downstream performance:

\begin{table*}[h]
\centering
\small
\setlength{\tabcolsep}{8pt}
\renewcommand{\arraystretch}{1.05}
\begin{tabular}{l|cc|cc|cc}
\hline
\textbf{Model} & \multicolumn{2}{c|}{\textbf{Math-500}} & \multicolumn{2}{c|}{\textbf{AIME-2025}} & \multicolumn{2}{c}{\textbf{GPQA}} \\
\cline{2-7}
 & Uniform & Adjusted & Uniform & Adjusted & Uniform & Adjusted \\
\hline
Nemotron-Elastic-6B  & 96.40 & 96.50 & 67.71 & 68.13 & 55.30 & 53.78 \\
Nemotron-Elastic-9B  & 97.40 & 97.25 & 75.00 & 75.42 & 62.75 & 62.50 \\
Nemotron-Elastic-12B & 97.33 & 97.70 & 72.29 & 75.83 & 61.11 & 63.25 \\
\hline
NanoV2-9B   & \multicolumn{2}{c|}{97.30} & \multicolumn{2}{c|}{71.43} & \multicolumn{2}{c}{63.01} \\
NanoV2-12B  & \multicolumn{2}{c|}{97.50} & \multicolumn{2}{c|}{72.50} & \multicolumn{2}{c}{65.28} \\
\hline
\end{tabular}
\caption{Budget sampling ablation. Adjusted non-uniform sampling (0.5, 0.3, 0.2 for 12B, 9B, 6B) achieves better balance across model sizes, particularly improving 12B accuracy on challenging benchmarks (AIME-2025: +3.54\%, GPQA: +2.14\%) compared to uniform sampling.}
\label{tab:sampling-ablation}
\end{table*}

The ablation demonstrates that adjusted sampling substantially improves performance for the full-budget model. For instance, on AIME-2025, the 12B model gains 3.54 percentage points with adjusted sampling, while maintaining competitive performance on other budgets. This suggests that multi-budget training requires careful load balancing to prevent negative transfer between budget targets.

\paragraph{Impact of Budget Sampling Strategy.}
We investigate the effect of budget sampling distribution through an ablation study comparing uniform budget allocation against our adjusted non-uniform sampling (Table~\ref{tab:sampling-ablation}). Results demonstrate that uniform sampling leads to performance imbalance during extended-context training: the 12B model's accuracy degrades significantly on challenging benchmarks, while smaller variants remain competitive. Our adjusted weighting ($p(\text{12B}) = 0.5, p(\text{9B}) = 0.3, p(\text{6B}) = 0.2$) recovers full-model performance by prioritizing gradients toward the largest variant, where reasoning capability demands greater architectural sophistication. The 12B model shows substantial improvements across multiple benchmarks, while smaller variants maintain stable performance. This ablation confirms that budget-aware curriculum design is essential for balanced multi-target elastic compression, and that Nemotron-Elastic-12B achieves competitive performance relative to baseline NanoV2 models while enabling zero-shot deployment across all budget variants.

\section{Related Work}

\textbf{Model Compression and Pruning.} Structured pruning has emerged as a powerful technique for LLM compression~\cite{ma2023llm, ashkboos2024slicegpt, xia2023sheared}. Recent work combines pruning with knowledge distillation for accuracy recovery~\cite{muralidharan2024compact}, achieving strong results on Transformer models.
Group-aware SSM pruning preserves structural constraints critical for sequence modeling while enabling hybrid model compression~\cite{taghibakhshi2025efficient}.
Unfortunately, these approaches require separate distillation for each target size.

\textbf{Hybrid SSM-Transformer Models.} Hybrid architectures combining Transformers with SSMs have shown promise for efficient long-context modeling~\cite{gu2023mamba, dao2024transformers, lieber2024jamba, glorioso2024zamba, blakeman2025nemotron}. Nemotron-H~\cite{blakeman2025nemotron} replaces 92\% of attention layers with Mamba2 blocks, achieving 3$\times$ inference speedup. Concurrent compression work~\cite{taghibakhshi2025minitron} introduces group-aware Mamba pruning but requires separate distillation per model size. 

\textbf{Elastic and Nested Architectures.} MatFormer~\cite{kudugunta2023matformer} and Flextron~\cite{cai2024flextron} pioneered nested weight-sharing for Transformers, training multiple sub-networks simultaneously. Extending the MatFormer methodology to SSMs, MatMamba~\cite{shukla2024matmambamatryoshkastatespace} introduces Matryoshka-style sub-block architecture for Mamba layers. MatFormer introduces MixnMatch heuristics for sub-network selection, while Flextron adds
input-adaptive routing for attention and MLP dimensions. However, neither supports: (1) hybrid Mamba-Attention architectures, (2) reasoning-focused two-stage training with extended context, or (3) heterogeneous layer-wise architecture selection via end-to-end learned routing. Google's Gemma 3n~\cite{gemma3n} recently demonstrated MatFormer-style nested models with conditional parameter loading, validating the practical deployment value of elastic architectures. Our work extends these foundations to reasoning models and hybrid architectures.

\textbf{Reasoning Model Training.} Reasoning-capable LLMs generate extended thought chains for complex problem-solving~\cite{wei2022chain, yao2023tree}, requiring long-context support for intermediate steps. Prior work on reasoning model optimization focuses on prompting strategies or reinforcement learning from reasoning traces~\cite{lightman2023lets}, but does not address architectural efficiency or elastic deployment. We demonstrate that reasoning models have fundamentally different training requirements—specifically, extended-context training (49K tokens) is \emph{critical} for maintaining reasoning performance in compressed variants, a requirement not present in standard LLM compression.

\section*{Conclusions}

This paper has presented Nemotron Elastic, the first elastic training framework for reasoning-capable LLMs. We demonstrate that elastic compression of reasoning models requires fundamentally different approaches than standard LLM compression, with extended-context training playing a critical role in preserving reasoning performance across deployment scales. 
Nemotron Elastic achieves strong results: deriving a model family from a single 12B parent requires only 110B training tokens—a 360x reduction versus training from scratch and a 7x reduction compared to sequential compression. This efficiency is achieved without compromising accuracy or introducing memory overhead, and having a constant memory footprint at deployment against the number of models in the family. During deployment, all the nested submodels can be extracted from the biggest model using zero-shot slicing. Our approach makes elastic reasoning model training practical for organizations with modest computational budgets.
Future directions for this work include scaling to larger model families, task-specific architecture selection, dynamic inference-time routing, and integration with quantization for extreme parameter reduction.

\section{Acknowledgments}
We would like to thank our colleagues and leaders at NVIDIA for their valuable input and support, including Akhiad Bercovich, Alex Fit-Florea, Joey Conway, Jonah Alben, Jonathan Cohen, Luis Vega, Michael Lightstone, Nave Assaf, Oleksii Kuchaiev, Ran Zilberstein, Terry Kong, Udi Karpas, and Zijia Chen.

\clearpage
{
  \small
  \bibliographystyle{unsrt}
  \bibliography{paper}
}

\end{document}